\documentclass[conference,10pt,a4paper]{IEEEtran}

\usepackage[noadjust]{cite}

\usepackage{graphicx}
\graphicspath{{figures/}}
\DeclareGraphicsExtensions{.pdf,.jpg,.png}

\usepackage{amsmath}
\usepackage{icomma}
\usepackage{xfrac}

\usepackage{array}

\usepackage{tikz}
\usepackage{csquotes}
\usepackage{booktabs}

\usepackage{pgfplots}

\usepackage[caption=false,font=footnotesize]{subfig}

\usepackage{url}

\definecolor{xblue}{RGB}{210,224,255}
\definecolor{xyellow}{RGB}{255,255,205}
\definecolor{xred}{RGB}{255,205,205}
\definecolor{xgreen}{RGB}{205,255,205}

\definecolor{cbgreen}{RGB}{127,201,127}
\definecolor{cborange}{RGB}{253,192,134}
\definecolor{cbviolet}{RGB}{190,174,212}

\pgfplotsset{minor grid style={dotted,white!75!black}}

\begin{document}
%
\title{Facial Expression Recognition using\\Convolutional Neural Networks: State of the Art}

\author{\IEEEauthorblockN{Christopher Pramerdorfer, Martin Kampel}
\IEEEauthorblockA{Computer Vision Lab, TU Wien\\
Vienna, Austria\\
Email: \{cpramer,kampel\}@caa.tuwien.ac.at}
}

\maketitle

\begin{abstract}
The ability to recognize facial expressions automatically enables novel applications in human-computer interaction and other areas. Consequently, there has been active research in this field, with several recent works utilizing Convolutional Neural Networks (CNNs) for feature extraction and inference. These works differ significantly in terms of CNN architectures and other factors. Based on the reported results alone, the performance impact of these factors is unclear. In this paper, we review the state of the art in image-based facial expression recognition using CNNs and highlight algorithmic differences and their performance impact. On this basis, we identify existing bottlenecks and consequently directions for advancing this research field. Furthermore, we demonstrate that overcoming one of these bottlenecks -- the comparatively basic architectures of the CNNs utilized in this field -- leads to a substantial performance increase. By forming an ensemble of modern deep CNNs, we obtain a FER2013 test accuracy of 75.2\%, outperforming previous works without requiring auxiliary training data or face registration.

\end{abstract}

%
\IEEEpeerreviewmaketitle


\section{Introduction}
\label{sec:introduction}


Being able to recognize facial expressions is key to nonverbal communication between humans, and the production, perception, and interpretation of facial expressions have been widely studied \cite{sariyanidi15}. Due to the important role of facial expressions in human interaction, the ability to perform \emph{Facial Expression Recognition} (FER) automatically via computer vision enables a range of novel applications in fields such as human-computer interaction and data analytics \cite{martinez16}.

\begin{figure}[!t]
\centering
\includegraphics[width=0.87\linewidth]{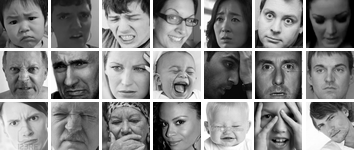}
\caption{Example images from the FER2013 dataset \cite{goodfellow15}, illustrating variabilities in illumination, age, pose, expression intensity, and occlusions that occur under realistic conditions. Images in the same column depict identical expressions, namely anger, disgust, fear, happiness, sadness, surprise, as well as neutral.}
\label{fig:fer2013-faces}
\end{figure}

Consequently, FER has been widely studied and significant progress has been made in this field. In fact, recognizing basic expressions under controlled conditions (e.g.~frontal faces and posed expressions) can now be considered a solved problem \cite{sariyanidi15}. The term \emph{basic expression} refers to a set of expressions that convey universal emotions, usually anger, disgust, fear, happiness, sadness, and surprise. Recognizing such expressions under naturalistic conditions is, however, more challenging. This is due to variations in head pose and illumination, occlusions, and the fact that unposed expressions are often subtle, as Fig.~\ref{fig:fer2013-faces} illustrates. Reliable FER under naturalistic conditions is mandatory in the aforementioned applications, yet still an unsolved problem \cite{sariyanidi15,martinez16}.

\emph{Convolutional Neural Networks} (CNNs) have the potential to overcome these challenges. CNNs have enabled significant performance improvements in related tasks (e.g.~\cite{krizhevsky12,he15,schroff15}), and several recent works on FER successfully utilize CNNs for feature extraction and inference (e.g.~\cite{tang13,yu15,kim16cvpr}). These works differ significantly in terms of CNN architecture, preprocessing, as well as training and test protocols, factors that all affect performance. It is therefore not possible to assess the impact of the CNN architecture and other factors based on the reported results alone. Being able to do so is, however, required in order to be able to identify existing bottlenecks in CNN-based FER, and consequently for improving FER performance.


The aim of this paper is to shed light on this matter by reviewing existing CNN-based FER methods and  highlighting their differences (Section \ref{sec:state_of_the_art}), as well as comparing the utilized CNN architectures empirically under consistent settings (Section \ref{sec:empirical_comparison}). On this basis, we identify existing bottlenecks and directions for improving FER performance. Finally, we confirm empirically that overcoming one such bottleneck improves performance substantially, demonstrating that modern deep CNNs achieve competitive results without auxiliary data or face registration (Section \ref{sec:Deep CNNs for FER}). An ensemble of such CNNs obtains a FER2013 \cite{goodfellow15} test accuracy of $75.2\%$, outperforming existing CNN-based FER methods.


In this paper, we consider the task of predicting basic expressions from single images using CNNs. For more general surveys, we refer to \cite{sariyanidi15,martinez16}. We note that it is straight-forward to adapt image-based methods to support image sequences by integrating per-frame results using graphical models. The conclusions drawn in this paper are thus relevant for sequence-based FER as well.


\section{State of the Art in CNN-Based FER}
\label{sec:state_of_the_art}


We review six state-of-the-art methods for CNN-based FER, highlight methodological differences, and discuss the reported performances. Most of these methods were evaluated on several databases in the original papers, the most common dataset being FER2013 \cite{goodfellow15}. For consistency, we study the methods as they were used for this dataset, and summarize and discuss the reported performances on this dataset.

FER2013 is a large, publicly available FER dataset consisting of 35,887 face crops. The dataset is challenging as the depicted faces vary significantly in terms of person age, face pose, and other factors (Fig.~\ref{fig:fer2013-faces}), reflecting realistic conditions. The dataset is split into training, validation, and test sets with 28,709, 3,589, and 3,589 samples, respectively. Basic expression labels are provided for all samples. All images are grayscale and have a resolution of 48 by 48 pixels. The human accuracy on this dataset is around 65.5\% \cite{goodfellow15}.

\begin{figure}[t]
\centering
\begin{tikzpicture}[scale=1.1, every node/.style={scale=1.1}]
%
%
\node at (0,0) {\includegraphics[width=4cm]{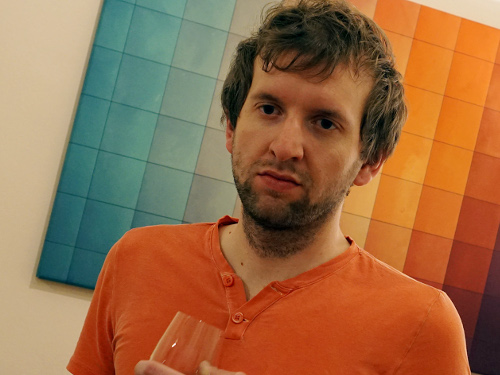}};
\draw[green,very thick] (-0.25,-0.35) rectangle (0.95,0.85);
\pgfmathsetmacro{\CEPS}{0.05}
\pgfmathsetmacro{\CX}{0.08}
\pgfmathsetmacro{\CY}{0.1}
\draw[red,thick] (\CX-\CEPS,\CY) -- (\CX+\CEPS,\CY);
\draw[red,thick] (\CX,\CY-\CEPS) -- (\CX,\CY+\CEPS);
\pgfmathsetmacro{\CX}{0.39}
\pgfmathsetmacro{\CY}{0.02}
\draw[red,thick] (\CX-\CEPS,\CY) -- (\CX+\CEPS,\CY);
\draw[red,thick] (\CX,\CY-\CEPS) -- (\CX,\CY+\CEPS);
\pgfmathsetmacro{\CX}{0.25}
\pgfmathsetmacro{\CY}{0.27}
\draw[red,thick] (\CX-\CEPS,\CY) -- (\CX+\CEPS,\CY);
\draw[red,thick] (\CX,\CY-\CEPS) -- (\CX,\CY+\CEPS);
\pgfmathsetmacro{\CX}{0.15}
\pgfmathsetmacro{\CY}{0.62}
\draw[red,thick] (\CX-\CEPS,\CY) -- (\CX+\CEPS,\CY);
\draw[red,thick] (\CX,\CY-\CEPS) -- (\CX,\CY+\CEPS);
\pgfmathsetmacro{\CX}{0.6}
\pgfmathsetmacro{\CY}{0.5}
\draw[red,thick] (\CX-\CEPS,\CY) -- (\CX+\CEPS,\CY);
\draw[red,thick] (\CX,\CY-\CEPS) -- (\CX,\CY+\CEPS);
\node at (2.5, 0) {$\boldsymbol{\Longrightarrow}$};
%
%
\node at (4.22,0) {\includegraphics[width=2.5cm]{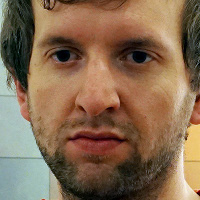}};
\pgfmathsetmacro{\CEPS}{0.08}
\pgfmathsetmacro{\CX}{3.75}
\pgfmathsetmacro{\CY}{0.55}
\draw[red,thick] (\CX-\CEPS,\CY) -- (\CX+\CEPS,\CY);
\draw[red,thick] (\CX,\CY-\CEPS) -- (\CX,\CY+\CEPS);
\pgfmathsetmacro{\CX}{4.7}
\pgfmathsetmacro{\CY}{0.53}
\draw[red,thick] (\CX-\CEPS,\CY) -- (\CX+\CEPS,\CY);
\draw[red,thick] (\CX,\CY-\CEPS) -- (\CX,\CY+\CEPS);
\pgfmathsetmacro{\CX}{4.13}
\pgfmathsetmacro{\CY}{-0.05}
\draw[red,thick] (\CX-\CEPS,\CY) -- (\CX+\CEPS,\CY);
\draw[red,thick] (\CX,\CY-\CEPS) -- (\CX,\CY+\CEPS);
\pgfmathsetmacro{\CX}{3.85}
\pgfmathsetmacro{\CY}{-0.48}
\draw[red,thick] (\CX-\CEPS,\CY) -- (\CX+\CEPS,\CY);
\draw[red,thick] (\CX,\CY-\CEPS) -- (\CX,\CY+\CEPS);
\pgfmathsetmacro{\CX}{4.5}
\pgfmathsetmacro{\CY}{-0.5}
\draw[red,thick] (\CX-\CEPS,\CY) -- (\CX+\CEPS,\CY);
\draw[red,thick] (\CX,\CY-\CEPS) -- (\CX,\CY+\CEPS);
\pgfmathsetmacro{\NOSEX}{4.17}
\pgfmathsetmacro{\NOSEY}{-0.08}
\draw[blue,thick] (\NOSEX,\NOSEY) circle (\CEPS);
\draw[blue,thick] (\NOSEX-0.28,\NOSEY-0.38) circle (\CEPS);
\draw[blue,thick] (\NOSEX+0.28,\NOSEY-0.38) circle (\CEPS);
\draw[blue,thick] (\NOSEX-0.43,\NOSEY+0.6) circle (\CEPS);
\draw[blue,thick] (\NOSEX+0.43,\NOSEY+0.6) circle (\CEPS);
\end{tikzpicture}
\caption{Illustration of a standard preprocessing pipeline, which involves face detection (green square), facial landmark detection (red crosses), registration to reference landmarks (blue circles), and illumination correction.}
\label{fig:face-registration}
\end{figure}
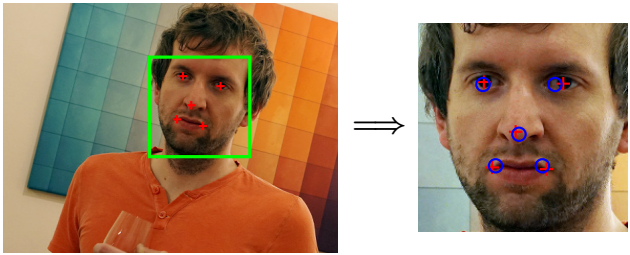

\subsection{Overview} 
\label{sub:sota_overview}

Image-based FER under naturalistic conditions has been an active research field for years, and several public challenges have been held to promote progress in this field. One such challenge was FER2013 \cite{goodfellow15}, which was won by one of the first CNN-based FER methods \cite{tang13}. The method uses an ensemble of CNNs trained to minimize the squared hinge loss.

In a more recent work, Yu and Zhang \cite{yu15} also utilize an ensemble of CNNs, and employ data augmentation at both training and test time in order to improve performance. Instead of performing ensemble voting via uniform averaging as in \cite{tang13}, ensemble predictions are integrated via weighted averaging with learned weights. The method ranked second in the recent EmotiW2015 challenge \cite{dhall15}.

The winner \cite{kim16} of this challenge employs a large committee of CNNs. Certain properties of the individual networks (e.g.\ input preprocessing and receptive field size) vary in order to obtain more diverse models. The ensemble predictions are integrated in a hierarchical fashion, with network weights assigned according to validation set performance.

Mollahosseini et al. \cite{mollahosseini15} trained a single CNN based on the Inception architecture \cite{szegedy15} on data compiled from multiple posed and naturalistic datasets in an effort to obtain a model that generalizes well across datasets.

In \cite{zhang2015} Zhang et al.\ present a method for inferring social relation traits from images using a Siamese network. In order to increase the amount of available training data, they utilize multiple datasets with heterogeneous labels. The authors present a patch-based registration and feature extraction technique, and perform feature integration via early fusion.

A recent work by Kim et al.\ \cite{kim16cvpr} shows that it is beneficial to use both unregistered and registered versions of a given face image during both training and testing. In order to prevent registration errors from affecting FER performance, registration is performed selectively based on the results of facial landmark detection. The authors show that registration can also be performed by deep networks, and that utilizing pose information captured by such networks leads to a small increase in FER performance (about 0.4\%).

\subsection{Methodological Differences} 
\label{sub:methodological_differences}

In order to highlight the methodological differences between these works, we break down each method into the three components (i) preprocessing, (ii) CNN architecture, and (iii) CNN training and inference.

\subsubsection{Preprocessing} 
\label{ssub:preprocessing}

Preprocessing entails operations that are applied once to each image. This typically includes face detection, face registration to compensate for pose variations, and means for correcting for illumination variations. Fig.~\ref{fig:face-registration} illustrates these steps.

Table \ref{tbl:sota_preprocessing} summarizes the preprocessing steps of every method. Only \cite{kim16} and \cite{yu15} perform face detection; all other methods rely on face crops provided by the datasets.

\begin{table}[t]
\centering
\caption{Preprocessing operations. FD: face detection, LM: facial landmark extraction., histeq: histogram equalization, lpf: linear plane fitting.}
\label{tbl:sota_preprocessing}
\begin{tabular}{ccccc}
    \toprule
    Method & FD & LM & Registration & Illumination \\ \midrule
    \cite{tang13} & no & no & no & normalize \\ 
    \cite{kim16} & several & \cite{xiong13} & rigid (LM) & several \\ 
    \cite{yu15} & several & no & no & histeq, lpf \\ 
    \cite{mollahosseini15} & no & \cite{xiong13} & affine (LM) & no \\ 
    \cite{zhang2015} & no & \cite{xiong13} & indirect & no \\ 
    \cite{kim16cvpr} & no & \cite{xiong13} & rigid (LM) & several \\ 
    \bottomrule
\end{tabular}
\end{table}

Face registration is common, with rigid or affine transformations based on extracted facial landmark locations being the most popular approach. This form of registration has the potential to improve FER performance \cite{kim16cvpr} provided that landmarks can be detected reliably. However, this is not always the case in practice due to challenging face poses and/or partial occlusions \cite{kim16cvpr}. There are different approaches to account for this problem; \cite{kim16} perform landmark detection on multiple versions of a given face image and utilize the detections with the highest detector confidence. \cite{kim16cvpr} perform alignment only if this confidence exceeds a threshold.

The majority of existing methods uses some form of illumination correction. In \cite{tang13} every image is normalized to have a mean of 0 and a norm of 100. \cite{yu15} employ histogram equalization and linear plane fitting. In \cite{kim16} and \cite{kim16cvpr} the methods vary between individual CNNs in the ensembles.

\subsubsection{CNN Architecture} 
\label{ssub:cnn_architecture}

Table \ref{tbl:sota_cnns} compares the utilized CNN architectures and their depths (number of layers with weights) and parameter counts. The counts were calculated assuming single-channel input images of size 48 by 48 pixels (the size of images in the FER2013 dataset). In some works, the CNNs operate on images of different sizes.

The table highlights that the individual architectures vary significantly in terms of layer composition, depth, and number of parameters. Most architectures are shallow compared to architectures in related fields \cite{parkhi15,szegedy15,he15}. As the corresponding papers lack details on architecture selection, the reason for this discrepancy is unknown.

The small size of available FER datasets such as FER2013 is not the limiting factor. First, deeper networks do not necessarily have more parameters, as shown in Table \ref{tbl:sota_cnns}. Second, deeper networks impose a stronger prior on the structure of the learned decision function, and this prior effectively combats overfitting \cite{goodfellow14}. Third, modern deep CNNs achieve impressive results on datasets with a similar size, such as CIFAR10 \cite{he15}. A possible explanation is that CNNs do not have to be as deep for FER; \cite{khorrami15} shows that a CNN with depth 5 is already able to learn discriminative high-level features. We postpone further discussions on this matter to Section \ref{sec:empirical_comparison}.

\begin{table}[t]
\centering
\caption{CNN architectures. C, P, N, I, and F stands for convolutional, pooling, response-normalization, inception, and fully connected layers, respectively.}
\label{tbl:sota_cnns}
\begin{tabular}{crrr}
    \toprule
    Method & Architecture & Depth & Parameters \\ \midrule
    \cite{tang13} & CPCPFF & 4 & 12.0 m \\ 
    \cite{kim16} & CPCPCPFF & 5 & 4.8 m \\ 
    \cite{yu15} & PCCPCCPCFFF & 8 & 6.2 m  \\ 
    \cite{mollahosseini15} & CPCPIIPIPFFF & 11 & 7.3 m \\
    \cite{zhang2015} & CPNCPNCPCFF & 6 & 21.3 m \\
    \cite{kim16cvpr} & CPCPCPFF & 5 & 2.4 m \\
    \bottomrule
\end{tabular}
\end{table}

\subsubsection{CNN Training and Inference} 
\label{ssub:cnn_training_inference}

Table \ref{tbl:sota_train_test} highlights the differences in terms of CNN training and inference. Of the six works compared, four use only the FER2013 training set for CNN training. \cite{mollahosseini15} and \cite{zhang2015} instead train on an union of seven and three datasets, respectively, in order to compensate for the fact that available FER image datasets are comparatively small.

\cite{zhang2015} and \cite{kim16cvpr} use additional features. In \cite{zhang2015} a vector of HoG features is computed from face patches and processed by the first fully connected layer of the CNN (early fusion). In \cite{kim16cvpr} the these features encode face pose information, and classifiers are trained to perform FER on this basis. Integration is performed via ensemble voting (late fusion).

All works except \cite{zhang2015} employ data augmentation during training in order to increase the amount of available data. Most works use standard augmentation methods that are not specific to FER, including horizontal mirroring and random cropping. The exception is \cite{kim16cvpr}, which additionally augments the training set by a registered version of every image.

In \cite{yu15} and \cite{kim16cvpr} augmentation is also performed at test time. In the former work, multiple perturbed versions of each test image are generated by applying affine transformations randomly, and the CNN output probabilities are averaged. The latter work follows the same approach but uses ten-crops (center and corner crops and mirrored versions) of a given image before and after face registration.

Most works use an ensemble of CNNs, whose predictions are integrated via different forms of averaging.

\begin{table}[t]
\centering
\caption{Differences in terms of CNN training and inference. AD: additional training data, AF: additional features, $+$: data augmentation, S,A: similarity/affine transform, T: translation, M: horizontal mirroring, reg: face registration.}
\label{tbl:sota_train_test}
\begin{tabular}{cccccc}
    \toprule
    Method & AD & AF & $+$~Train & $+$~Test & Ensemble \\ \midrule
    \cite{tang13} & no & no & S,M & -- & average \\ 
    \cite{kim16} & no & no & T,M & -- & hierarchy \\ 
    \cite{yu15} & no & no & A,M & A & weighted \\
    \cite{mollahosseini15} & yes & no & ten-crop & -- & -- \\
    \cite{zhang2015} & yes & yes & -- & -- & -- \\
    \cite{kim16cvpr} & no & yes & T,M,reg & ten-crop,reg & average \\
    \bottomrule
\end{tabular}
\end{table}

\subsection{Reported Results on FER2013} 
\label{sub:reported_results_on_fer2013}

\begin{figure}[t]
\begin{center}
\begin{tikzpicture}
\begin{axis}[
    ybar,
    ymajorgrids = true,
    yminorgrids = true,
    ymin = 66,
    ymax = 76,
    xtick = {1,...,6},
    ytick = {66,...,76},
    minor y tick num = 1,
    ylabel={Accuracy [\%]},
    legend pos=north west,
    every axis x label/.style={at={(ticklabel cs:0.5)},anchor=near ticklabel},
    every axis y label/.style={at={(ticklabel cs:0.5)},rotate=90,anchor=near ticklabel},
    xticklabels = {\cite{tang13}, \cite{kim16}, \cite{yu15}, \cite{mollahosseini15}, \cite{zhang2015}, \cite{kim16cvpr}},
    bar width = 14,
    width = 0.9\linewidth,
    height = 0.6\linewidth,
]
\addplot[fill=cbgreen] coordinates {(1,71.2) (2,72.7) (3,72.1) (4,66.4) (5,75.1) (6,73.7)};
\end{axis}
\end{tikzpicture}
\end{center}
\caption{Reported FER2013 test results.}
\label{fig:sota-perf-fer2013}
\end{figure}
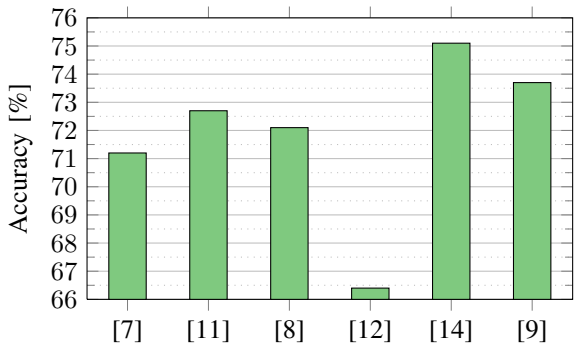

Fig.~\ref{fig:sota-perf-fer2013} compares the reported FER2013 test accuracies. Based on the methodological differences and these results, we draw the following remarks.

The three best-performing works use comparatively shallow CNNs (depths of 5 and 6). The work utilizing the deepest and most modern (in terms of layer types and arrangement) CNN \cite{mollahosseini15} performs worst on this dataset. However, a direct comparison of the results is not possible because the CNN was trained on a superset of FER2013; in case of \cite{mollahosseini15}, this presumably has a negative impact on FER2013 test performance. On the other hand, Zhang et al.\ \cite{zhang2015} demonstrate that utilizing additional training data in a way that accounts for dataset bias improves the performance on FER2013.

The three best-performing methods use face registration, suggesting that registration is beneficial even under challenging conditions (according to \cite{zhang2015}, facial landmark extraction is inaccurate for about 15\% of images in the FER dataset).

Data augmentation and ensemble voting are important in order to improve generalization performance. The best-performing work that was trained on FER2013 alone uses the most comprehensive form of data augmentation during both training and testing. Ensemble voting reportedly improves the test accuracy by 2-3\% \cite{yu15,kim16cvpr}.

\section{Empirical Comparison} 
\label{sec:empirical_comparison}

Im summary, Fig.~\ref{fig:sota-perf-fer2013} shows that the best-performing FER methods utilize CNNs that are shallow and basic (in terms of layer types and arrangement) compared to the state of the art in related fields \cite{schroff15,he15}, contradicting the general trend towards deeper and deeper networks.

This, however, does not mean that such CNNs are inapplicable for FER because the CNN architecture is one of many factors that influence FER performance. In order to obtain more information on this matter, we perform an empirical comparison of the utilized CNN architectures.

\subsection{Experiments} 
\label{sub:comparison_experiments}

We train and test all CNN architectures utilized in the compared works using the same protocols.

\subsubsection{CNN Architectures} 
\label{ssub:cnn_architectures_exp}

We test all architectures as they are described in the corresponding papers, apart from the following differences. We add batch normalization \cite{ioffe15} layers after every convolutional and fully connected (fc) layer for robustness to suboptimal network initialization. Furthermore, we add a dropout layer \cite{srivastava2014} after the first fc layer.

\cite{kim16} uses an ensemble of CNNs with different receptive fields and numbers of neurons in the first fc layer. We use the configuration with $3\times3$ receptive fields and 2,048 neurons.

\subsubsection{Dataset and Preprocessing} 
\label{ssub:dataset_and_preprocessing}

In order to enable comparison with reported results, we perform all experiments on the FER2013 dataset \cite{goodfellow15}, adhering to the official training, validation, and test sets. We use the face crops as provided by the dataset and employ histogram equalization for illumination correction. This follows subtraction and division by the mean and standard deviation over all training pixels, respectively.

We do not perform landmark-based registration for two reasons. First, this prevents registration errors from affecting the results. Second, this forces the CNNs to learn to compensate for pose variations, potentially leading to more general models.

\subsubsection{CNN Training and Inference} 
\label{ssub:cnn_training_and_inference}

We train every architecture for up to 300 epochs, optimizing the cross-entropy loss using stochastic gradient descent with a momentum of 0.9. The initial learning rate, batch size, and weight decay are fixed at 0.1,  128, and 0.0001, respectively. The learning rate is halved if the validation accuracy does not improve for 10 epochs.

For training data augmentation we use horizontal mirroring and random crops of size 48 by 48 pixels after zero-padding (as the input images are already of this size). We ensure that all CNNs see the same training samples (after augmentation) in the same order, thereby enabling a fair comparison.

As the individual architectures were designed for training on larger datasets and/or other forms of data augmentation, training using the same regularization parameters would be unfair. In order to account for this, we perform a grid-search to find out an optimal dropout rate for every architecture.

Finally, the best model obtained for each architecture in terms of validation accuracy is tested on the test set using standard ten-crop oversampling \cite{he15}.

\subsubsection{Feature Comparison} 
\label{ssub:feature_comparison}

Furthermore, we empirically compare the quality of the features learned by the models with the highest validation accuracy. This is accomplished by replacing the fc backends of the trained CNNs with a two-layer MLP with 1,024 hidden units, which is then trained using the above protocol. All other parameters are fixed, effectively causing the MLP to learn to perform FER using the features extracted by the pretrained frontend of the network.

Doing so allows a more direct comparison of the results because the backends of the resulting models are no longer different. With different backends, a more powerful backend could mask limitations in the learned representations. Fixing the backend enables us to study the impact of the network depth on the capabilities of the learned representations.

\subsection{Results and Discussion} 
\label{sub:results_and_discussion}

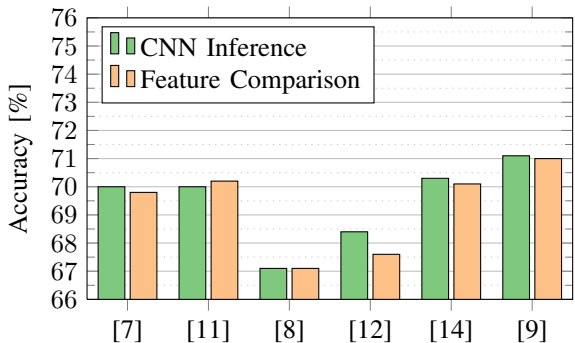
\begin{figure}[t]
\begin{center}
\begin{tikzpicture}
\begin{axis}[
    ybar,
    ymajorgrids = true,
    yminorgrids = true,
    ymin = 66,
    ymax = 76,
    xtick = {1,...,6},
    ytick = {66,...,76},
    minor y tick num = 1,
    ylabel={Accuracy [\%]},
    legend pos=north west,
    every axis x label/.style={at={(ticklabel cs:0.5)},anchor=near ticklabel},
    every axis y label/.style={at={(ticklabel cs:0.5)},rotate=90,anchor=near ticklabel},
    xticklabels = {\cite{tang13}, \cite{kim16}, \cite{yu15}, \cite{mollahosseini15}, \cite{zhang2015}, \cite{kim16cvpr}},
    bar width = 10,
    width = 0.9\linewidth,
    height = 0.6\linewidth,
    legend cell align=left
]
\addplot[fill=cbgreen] coordinates {(1,70.0) (2,70.0) (3,67.1) (4,68.4) (5,70.3) (6,71.1)};
\addplot[fill=cborange] coordinates {(1,69.8) (2,70.2) (3,67.1) (4,67.6) (5,70.1) (6,71.0)};
\legend{CNN Inference,Feature Comparison}
\end{axis}
\end{tikzpicture}
\end{center}
\caption{Ten-crop test results of the different network architectures (green), and when using pretrained frontends as feature extractors (orange).}
\label{fig:related-perf}
\end{figure}

Fig.~\ref{fig:related-perf} summarizes the results. All results, except those for \cite{mollahosseini15}, are lower than the reported ones (Fig.~\ref{fig:sota-perf-fer2013}). The main reason for this is the lack of ensemble voting, as most results are comparable to reported results of single networks \cite{yu15,kim16cvpr}. 

In some cases, the differences cannot be explained by the lack of ensemble voting; in case of \cite{zhang2015}, the reason for the lower performance is the lack of auxiliary training data. The reported accuracy in \cite{mollahosseini15} is lower than the measured accuracy. This is also explained by the additional training data, which in this case has a negative effect on FER2013 performance. This shows that auxiliary training data has the potential to significantly improve FER performance, provided that care is taken in order to address dataset bias.

In case of \cite{yu15}, the measured accuracy is about 3\% lower than the reported one using a single CNN. We tested both stochastic pooling (as used in \cite{yu15}) and max pooling, but in both cases were unable to reach the reported accuracy.

Overally, shallower CNN architectures again perform better than deeper ones (cf.\ Table \ref{tbl:sota_cnns}). This also applies to the learned features. This, however, does not confirm that modern deep networks are not suitable for FER; there is only one architecture in the comparison that qualifies as such \cite{mollahosseini15}, and some architectural choices of this network are questionable (initial convolution with a $7\times7$ receptive field, which appears too large given the input resolution, and a wide backend in the form of a three-layer MLP with 4,096 units in the first layer).

On the contrary, we postulate that modern deep networks can outperform the shallow architectures of current works, based on findings in related research fields \cite{schroff15,he15}. In Section \ref{sec:Deep CNNs for FER} we perform experiments to confirm this hypothesis.

\subsection{Current Bottlenecks} 
\label{sub:current_bottlenecks}

In this section, we highlight major bottlenecks in CNN-based FER based on the reported and measured results. We postulate that overcoming these bottlenecks will lead to substantial improvements in FER performance.

The CNN architectures used are basic and shallow compared to state-of-the-art architectures in related fields \cite{parkhi15,szegedy15,he15}.

Most works use general data augmentation techniques such as random crops and mirroring, which are not optimized for the task at hand. Kim et al.\ \cite{kim16cvpr} show that face-aware data augmentation via face registration improves performance. This approach can be extended, for instance by data augmentation via frontalized samples \cite{hassner15}, or by synthesizing faces in various poses via 3D face pose estimation \cite{taigman14,jeni2015dense}. Such data augmentation techniques, utilized for both training and inference, have the potential to effectively compensate for the limited size of current FER datasets.

For the same reason, training on a combination of multiple datasets can lead to significant improvements, provided that dataset bias is accounted for. \cite{zhang2015} show that this is beneficial even with  datasets with heterogeneous labels.

Lastly, we put forward that the biggest bottleneck that currently hinders FER performance is the fact that there is no publicly available dataset that is large by current deep learning standards. The introduction of datasets with hundreds of thousands or millions of images has enabled significant performance gains in related research fields such as face recognition \cite{taigman14,schroff15,parkhi15}. In contrast, FER2013, one of the largest FER image datasets available, has only 35,887 images. Compiling a large FER dataset is a laborious task due to the challenging annotation process; assigning correct expression labels in presence of subtle expressions, partial occlusions, and pose variations is a challenging task for humans \cite{sariyanidi15}.

\section{Deep CNNs for FER}
\label{sec:Deep CNNs for FER}

In this section, we confirm experimentally that overcoming one of these bottlenecks, the comparatively shallow and basic CNN architectures of current FER methods, leads to a substantial improvement in accuracy on FER2013.

\subsection{Experiments} 
\label{sub:experiments}

In order to enable a fair comparison with the results in Section \ref{sec:empirical_comparison}, we use the exact same dataset, preprocessing, as well as training and testing protocols. 

\subsubsection{CNN Architectures} 
\label{ssub:cnn_architectures}


We consider three CNNs whose architectures are summarized in Table \ref{tbl:deep-architectures}. All are inspired by current state-of-the-art architectures in related fields:

\textbf{VGG.} An architecture similar to VGG-B \cite{simonyan14} but with one CCP block less. We also use dropout after each such block (this improved the validation accuracy by around $1\%$). For consistency with Section\ \ref{sec:empirical_comparison}, the backend consists of a single hidden layer with 1024 units.

\textbf{Inception.} An architecture similar to GoogLeNet \cite{szegedy15}, but with a more consistent structure and without initial strided convolutions or pooling (the input images are already small enough). The net uses a consistent distribution of feature map sizes in a given Inception layer that is based on the number of $3\times3$ features maps $n$; the $1\times1, 3\times3$ reduce, $5\times5$ reduce, $5\times5$, and pool projection layers have \sfrac{3}{4}$n$, \sfrac{1}{2}$n$, \sfrac{1}{8}$n$, \sfrac{1}{4}$n$, and \sfrac{1}{4}$n$ feature maps, respectively. $n$ is initialized to $32$ and increased by $32$ after every Inception layer.

\textbf{ResNet.} Our architecture is identical to the 34-layer ResNet from \cite{he15}, but without the initial CP block. Our network is also more narrow, having 256 feature maps in the final residual group to reduce the number of parameters. We use dropout after the final pooling layer.

\begin{table}[t]
\centering
\caption{Tested deep architectures and their ten-crop test accuracy on FER2013. 3R means group of three residual blocks.}
\label{tbl:deep-architectures}
\begin{tabular}{lrccc}
    \toprule
    Name & Architecture & Depth & Parameters & Accuracy \\ \midrule
    VGG & CCPCCPCCPCCPFF & 10 & 1.8 m & 72.7\% \\
    Inception & CIPIIPIIPIIPF & 16 & 1.2 m & 71.6\% \\ 
    ResNet & 3R4R6R3RPF & 33 & 5.3 m & 72.4\% \\
    \bottomrule
\end{tabular}
\end{table}

VGG and Inception have less parameters than any of the architectures used in the pertinent literature, despite being significantly deeper (cf.~Tables \ref{tbl:deep-architectures}. and \ref{tbl:sota_cnns}). Even the very deep ResNet has fewer parameters than most of these architectures.

We did not specifically search for architectures that perform well on FER2013. Our goal is to confirm that modern deep architectures generally perform well, not to obtain the absolute best accuracies on this dataset.

\subsubsection{CNN Ensembles} 
\label{ssub:cnn_ensembles}

In order to demonstrate the potential of an ensemble of such deep CNNs, we perform an exhaustive search to identify optimal ensembles of up to 8 models in terms of FER2013 validation accuracy.

\subsection{Results and Discussion} 
\label{sub:results_and_discussion_own}

The test accuracies of the individual models with the best validation accuracies are given in Table \ref{tbl:deep-architectures}. The best modern deep model outperforms the best shallow model by almost 2\% under identical conditions (cf.\ Fig.~\ref{fig:related-perf}). All considered architectures outperform the best shallow model, including Inception, which has only half as many parameters. These results confirm that utilizing modern deep architectures has the potential to substantially improve FER performance.

Our individual CNNs already perform competitively to previous works that utilize ensemble voting (Fig.~\ref{fig:sota-perf-fer2013}). By forming an ensemble of 8 such CNNs, we achieve a FER2013 test accuracy of 75.2\%, performing comparably to the current best method we are aware of \cite{zhang2015}.

Our ensemble of deep models obtains state-of-the-art performance without utilizing additional training data or features, comprehensive data augmentation, or requiring face registration. By not requiring face registration, our FER method is conceptually simpler than previous methods and not affected by registration errors. We expect that utilizing auxiliary training data and comprehensive, FER-specific data augmentation would improve the performance further.

This paper has been studying the FER performance by means of the FER2013 dataset, which is the most common image dataset in CNN-based FER and one of the largest publicly available datasets in this field. (There are several video datasets that contain a much larger number of frames, but these frames are naturally highly correlated and the number of subjects in such datasets is small.) Still, results obtained on this and other FER datasets are only indicative of real-world FER performance due to dataset bias. This limitation applies not only to this study but to FER research in general.

\section{Conclusions}


In this paper we have reviewed the state of the art in CNN-based FER, highlighted key differences between the individual works, and compared and discussed their performance with a focus on the underling CNN architectures. On this basis, we have identified existing bottlenecks and consequently means for advancing the state of the art in this challenging research field. Furthermore, we have shown that overcoming one such bottleneck by employing modern deep CNNs leads to a significant improvement in FER2013 performance. Finally, we have demonstrated that an ensemble of such CNNs outperforms state of the art methods without the use of additional training data or requiring face registration.


We expect that overcoming the remaining bottlenecks identified in this paper will result in further substantial performance improvements. For the future, we plan to investigate ways for overcoming these bottlenecks, with a focus on FER-specific data augmentation. Furthermore, we will study the bias that affects FER2013 and other datasets, and investigate the possibility of creating a new, more comprehensive, and publicly available FER dataset.


\bibliographystyle{IEEEtran}
\bibliography{IEEEabrv,literature}

\begin{thebibliography}{10}
\providecommand{\url}[1]{#1}
\csname url@samestyle\endcsname
\providecommand{\newblock}{\relax}
\providecommand{\bibinfo}[2]{#2}
\providecommand{\BIBentrySTDinterwordspacing}{\spaceskip=0pt\relax}
\providecommand{\BIBentryALTinterwordstretchfactor}{4}
\providecommand{\BIBentryALTinterwordspacing}{\spaceskip=\fontdimen2\font plus
\BIBentryALTinterwordstretchfactor\fontdimen3\font minus
  \fontdimen4\font\relax}
\providecommand{\BIBforeignlanguage}[2]{{%
\expandafter\ifx\csname l@#1\endcsname\relax
\typeout{** WARNING: IEEEtran.bst: No hyphenation pattern has been}%
\typeout{** loaded for the language `#1'. Using the pattern for}%
\typeout{** the default language instead.}%
\else
\language=\csname l@#1\endcsname
\fi
#2}}
\providecommand{\BIBdecl}{\relax}
\BIBdecl

\bibitem{sariyanidi15}
E.~Sariyanidi, H.~Gunes, and A.~Cavallaro, ``{Automatic analysis of facial
  affect: A survey of registration, representation, and recognition},''
  \emph{IEEE Transactions on Pattern Analysis and Machine Intelligence (PAMI)},
  vol.~37, no.~6, pp. 1113--1133, 2015.

\bibitem{martinez16}
M.~V. B.~Martinez, ``{Advances, Challenges, and Opportunities in Automatic
  Facial Expression Recognition},'' in \emph{Advances in Face Detection and
  Facial Image Analysis}.\hskip 1em plus 0.5em minus 0.4em\relax Springer,
  2016, pp. 63--100.

\bibitem{goodfellow15}
I.~J. Goodfellow, D.~Erhan, P.~L. Carrier, A.~Courville, M.~Mirza, B.~Hamner,
  W.~Cukierski, Y.~Tang, D.~Thaler, D.-H. Lee, Y.~Zhou, C.~Ramaiah, F.~Feng,
  R.~Li, X.~Wang, D.~Athanasakis, J.~Shawe-Taylor, M.~Milakov, J.~Park,
  R.~Ionescu, M.~Popescu, C.~Grozea, J.~Bergstra, J.~Xie, L.~Romaszko, B.~Xu,
  Z.~Chuang, and Y.~Bengio, ``{Challenges in representation learning: A report
  on three machine learning contests},'' \emph{Neural Networks}, vol.~64, pp.
  59--63, 2015.

\bibitem{krizhevsky12}
A.~Krizhevsky, I.~Sutskever, and G.~E. Hinton, ``{ImageNet Classification with
  Deep Convolutional Neural Networks},'' in \emph{Advances in Neural
  Information Processing Systems (NIPS)}, 2012, pp. 1097--1105.

\bibitem{he15}
K.~He, X.~Zhang, haoqing Ren, and J.~Sun, ``{Deep Residual Learning for Image
  Recognition},'' \emph{CoRR}, vol. 1512, 2015.

\bibitem{schroff15}
F.~Schroff, D.~Kalenichenko, and J.~Philbin, ``{FaceNet: A Unified Embedding
  for Face Recognition and Clustering},'' in \emph{IEEE Conference on Computer
  Vision and Pattern Recognition (CVPR)}, 2015.

\bibitem{tang13}
Y.~Tang, ``{Deep Learning using Support Vector Machines},'' in
  \emph{International Conference on Machine Learning (ICML) Workshops}, 2013.

\bibitem{yu15}
Z.~Yu and C.~Zhang, ``{Image based static facial expression recognition with
  multiple deep network learning},'' in \emph{ACM International Conference on
  Multimodal Interaction (MMI)}, 2015, pp. 435--442.

\bibitem{kim16cvpr}
B.-K. Kim, S.-Y. Dong, J.~Roh, G.~Kim, and S.-Y. Lee, ``{Fusing Aligned and
  Non-Aligned Face Information for Automatic Affect Recognition in the Wild: A
  Deep Learning Approach},'' in \emph{IEEE Conf. Computer Vision and Pattern
  Recognition (CVPR) Workshops}, 2016, pp. 48--57.

\bibitem{dhall15}
A.~Dhall, O.~Ramana~Murthy, R.~Goecke, J.~Joshi, and T.~Gedeon, ``{Video and
  Image Based Emotion Recognition Challenges in the Wild: EmotiW 2015},'' in
  \emph{ACM International Conference on Multimodal Interaction (ICMI)}, 2015,
  pp. 423--426.

\bibitem{kim16}
B.-K. Kim, J.~Roh, S.-Y. Dong, and S.-Y. Lee, ``{Hierarchical committee of deep
  convolutional neural networks for robust facial expression recognition},''
  \emph{Journal on Multimodal User Interfaces}, pp. 1--17, 2016.

\bibitem{mollahosseini15}
A.~Mollahosseini, D.~Chan, and M.~H. Mahoor, ``{Going Deeper in Facial
  Expression Recognition using Deep Neural Networks},'' \emph{CoRR}, vol. 1511,
  2015.

\bibitem{szegedy15}
C.~Szegedy, W.~Liu, Y.~Jia, P.~Sermanet, S.~Reed, D.~Anguelov, D.~Erhan,
  V.~Vanhoucke, and A.~Rabinovich, ``{Going Deeper with Convolutions},'' in
  \emph{IEEE Conference on Computer Vision and Pattern Recognition (CVPR)},
  2015.

\bibitem{zhang2015}
Z.~Zhang, P.~Luo, C.-C. Loy, and X.~Tang, ``{Learning Social Relation Traits
  from Face Images},'' in \emph{Proc. IEEE Int. Conference on Computer Vision
  (ICCV)}, 2015, pp. 3631--3639.

\bibitem{xiong13}
X.~Xiong and F.~Torre, ``{Supervised descent method and its applications to
  face alignment},'' in \emph{IEEE Conference on Computer Vision and Pattern
  Recognition (CVPR)}, 2013, pp. 532--539.

\bibitem{parkhi15}
O.~M. Parkhi, A.~Vedaldi, and A.~Zisserman, ``Deep face recognition,'' in
  \emph{British Machine Vision Conference}, 2015.

\bibitem{goodfellow14}
I.~J. Goodfellow, Y.~Bulatov, J.~Ibarz, S.~Arnoud, and V.~D. Shet,
  ``{Multi-digit Number Recognition from Street View Imagery using Deep
  Convolutional Neural Networks},'' \emph{CoRR}, vol. 6082, 2013.

\bibitem{khorrami15}
P.~Khorrami, T.~Paine, and T.~Huang, ``{Do Deep Neural Networks Learn Facial
  Action Units When Doing Expression Recognition?}'' in \emph{IEEE
  International Conference on Computer Vision (ICCV) Workshops}, 2015, pp.
  19--27.

\bibitem{ioffe15}
S.~Ioffe and C.~Szegedy, ``{Batch Normalization: Accelerating Deep Network
  Training by Reducing Internal Covariate Shift},'' \emph{CoRR}, vol. 1502,
  2015.

\bibitem{srivastava2014}
N.~Srivastava, G.~Hinton, A.~Krizhevsky, I.~Sutskever, and R.~Salakhutdinov,
  ``{Dropout: A simple way to prevent neural networks from overfitting},''
  \emph{The Journal of Machine Learning Research}, vol.~15, no.~1, pp.
  1929--1958, 2014.

\bibitem{hassner15}
T.~Hassner, S.~Harel, E.~Paz, and R.~Enbar, ``{Effective Face Frontalization in
  Unconstrained Images},'' in \emph{IEEE Conf. Computer Vision and Pattern
  Recognition (CVPR)}, June 2015.

\bibitem{taigman14}
Y.~Taigman, M.~Yang, M.~Ranzato, and L.~Wolf, ``{Deepface: Closing the gap to
  human-level performance in face verification},'' in \emph{IEEE Conference on
  Computer Vision and Pattern Recognition (CVPR)}, 2014, pp. 1701--1708.

\bibitem{jeni2015dense}
L.~A. Jeni, J.~F. Cohn, and T.~Kanade, ``{Dense 3D Face Alignment from 2D
  Videos in Real-Time},'' in \emph{IEEE Int. Conf. Automatic Face and Gesture
  Recognition (FG)}, vol.~1.\hskip 1em plus 0.5em minus 0.4em\relax IEEE, 2015,
  pp. 1--8.

\bibitem{simonyan14}
K.~Simonyan and A.~Zisserman, ``{Very Deep Convolutional Networks for
  Large-Scale Image Recognition},'' \emph{CoRR}, vol. 1409, 2014.

\end{thebibliography}

\end{document}